\documentclass[10pt,twocolumn,letterpaper]{article}

\usepackage{iccv}
\usepackage{times}
\usepackage{epsfig}
\usepackage{graphicx}
\usepackage{amsmath}
\usepackage{amssymb}
\usepackage{subcaption}
\usepackage{subfloat} 
\usepackage{caption} 

\usepackage{multicol}
\usepackage{multirow}
\usepackage{array}
\usepackage{float}
\usepackage{stfloats}
\usepackage{booktabs}
\usepackage{amsthm}
\usepackage{amsthm}
\usepackage{mathrsfs}
\usepackage{makecell}
\usepackage{algpseudocode}
\usepackage{pifont}
\usepackage[frozencache,cachedir=./]{minted}

\usepackage[breaklinks=true,bookmarks=false]{hyperref}

\iccvfinalcopy 

\def\iccvPaperID{5328} 

\ificcvfinal\fi

\begin{document}

\title{Inherent Redundancy in Spiking Neural Networks}

\author{Man Yao$^{1,2,3*}$, Jiakui Hu$^{4,2}$\thanks{These authors contribute equally to this work}, Guangshe Zhao$^{1\dag}$, Yaoyuan Wang$^{5}$, Ziyang Zhang$^{5}$, Bo Xu$^{2}$, Guoqi Li$^{2}$\thanks{Corresponding author}\\
$^1$School of Automation Science and Engineering, Xi’an Jiaotong University, Xi’an, China \\
$^2$Institute of Automation, Chinese Academy of Sciences, Beijing, China\\
$^3$Peng Cheng Laboratory, Shenzhen, China \\
$^4$Peking University Health Science Center, Peking University, Beijing, China\\ 
$^5$Advanced Computing and Storage Lab, Huawei Technologies Co Ltd.\\
{\tt\small manyao@stu.xjtu.edu.cn, jkhu29@stu.pku.edu.cn, zhaogs@mail.xjtu.edu.cn, guoqi.li@ia.ac.cn}
}


\maketitle
\ificcvfinal\thispagestyle{empty}\fi

\newcommand{\rone}[1]{\textcolor{black}{#1}}
\renewcommand{\algorithmicrequire}{\textbf{Input:}}  
\renewcommand{\algorithmicensure}{\textbf{Output:}} %
\begin{abstract}
Spiking Neural Networks (SNNs) are well known as a promising energy-efficient alternative to conventional artificial neural networks. Subject to the preconceived impression that SNNs are sparse firing, the analysis and optimization of inherent redundancy in SNNs have been largely overlooked, thus the potential advantages of spike-based neuromorphic computing in accuracy and energy efficiency are interfered. In this work, we pose and focus on three key questions regarding the inherent redundancy in SNNs. We argue that the redundancy is induced by the spatio-temporal invariance of SNNs, which enhances the efficiency of parameter utilization but also invites lots of noise spikes. Further, we analyze the effect of spatio-temporal invariance on the spatio-temporal dynamics and spike firing of SNNs. Then, motivated by these analyses, we propose an Advance Spatial Attention (ASA) module to harness SNNs' redundancy, which can adaptively optimize their membrane potential distribution by a pair of individual spatial attention sub-modules. In this way, noise spike features are accurately regulated. Experimental results demonstrate that the proposed method can significantly drop the spike firing with better performance than state-of-the-art SNN baselines. Our code is available in \url{https://github.com/BICLab/ASA-SNN}.
\end{abstract}

\section{Introduction}\label{sec:intro}
\begin{figure}[!tbp]
\setlength{\belowcaptionskip}{-0.3cm}
\centering
\includegraphics[scale=0.55]{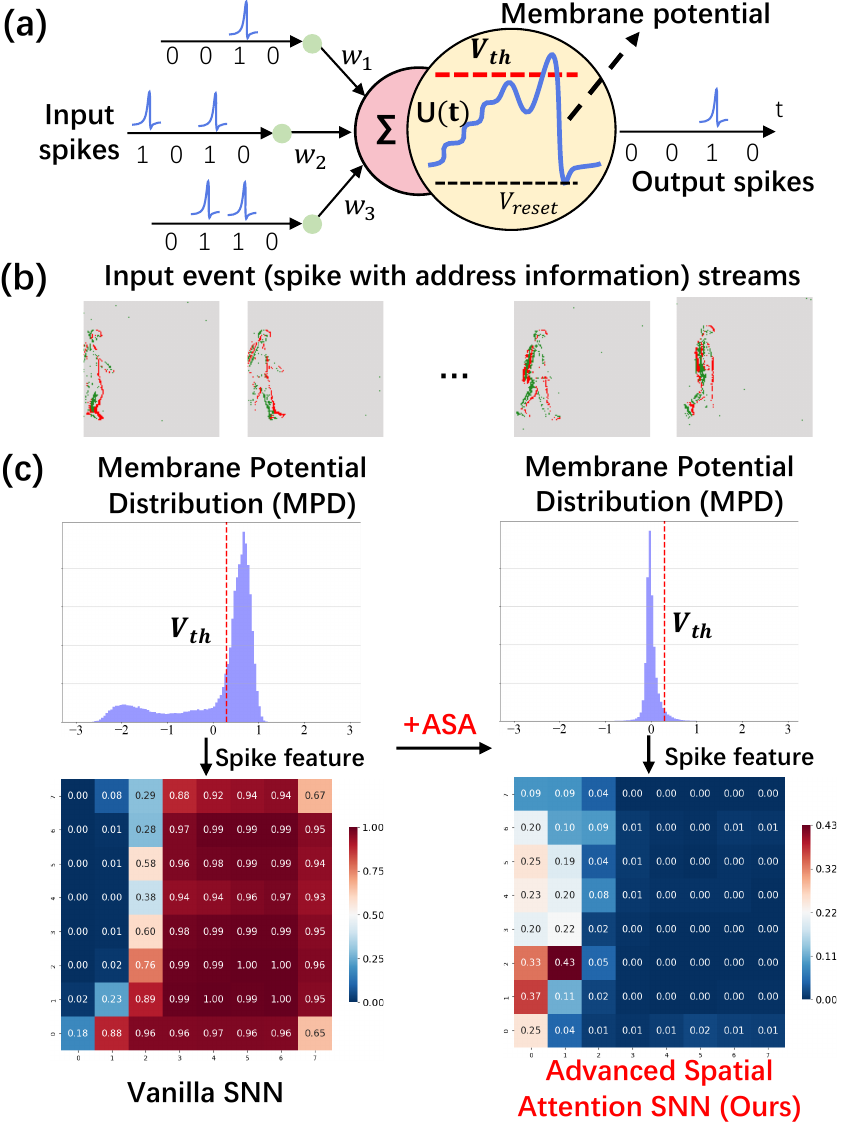}
\caption{(a) Spatio-temporal dynamics of spiking neurons with binary spike input and output, synaptic weight $w$, membrane potential $U(t)$, threshold $V_{th}$ and hard reset membrane potential $V_{reset}$. (b) An example of an event stream. (c) Examples of changes in the spike responses of vanilla SNN and ASA-SNN, in terms of Membrane Potential Distribution (MPD) and spike feature. Each pixel value on the spike feature represents the firing rate of a neuron. Noise spike feature fires lots of spikes while concentrating on insignificant background information (large area red). The ASA module can shift the spike pattern of SNNs to drop spike counts by optimizing the MPD.}
\label{fig:overview_of_paper}
\end{figure}

\rone{By mimicking the spatio-temporal dynamics behaviors of biological neurons, Spiking Neural Networks (SNNs) pose a paradigm shift in information encoding and transmitting \cite{Nature_2,schuman2022opportunities}. Spiking neurons only fire when the membrane potential is greater than the threshold (Figure~\ref{fig:overview_of_paper}\textbf{a}), in theory, these complex internal dynamics make the representation ability to spiking neurons more powerful than existing artificial neurons \cite{Maass_1997_LIF}. Moreover, spike-based binary communication (0/1 spike) enables SNNs to be \emph{event-driven} when deployed on neuromorphic chips \cite{davies2018loihi,Nature_1}, i.e., performing cheap synaptic Accumulation (AC) and bypassing computations on zero inputs or activations \cite{eshraghian_2021_training_lesson,Deng_2020_rethink_ann_snn}. }

\rone{For a long time, when referring to spike-based neuromorphic computing, people naturally believe that its computation is sparse due to the event-driven feature. Subject to this preconceived impression, although it is generally agreed that sparse spike firing is the key to achieving high energy efficiency in neuromorphic computing, there is a lack of systematic and in-depth analysis of redundancy in SNNs. Existing explorations are limited to specific methods of dropping spike counts.} For instance, several algorithms have been proposed to exploit spike-aware sparsity regularization and compression by adding a penalty function\cite{deng_TNNLS_2021,yin_2021_sparse_activity_1,zenke_2021_sparse_activity_2,yin_2021_NMI,neil_2016_activity_3,kundu_2021_spike_thrift}, designing network structures with fewer spikes using neural architecture search techniques \cite{AutoSNN_2022,kim2022neural}, or developing data-dependent models to regulate spike firing based on the input data \cite{yao_2021_TASNN,yao2022attention}. \rone{Generally, employing these methods to reduce spikes incurs a loss of accuracy or significant additional computation.}

In this work, we provide a novel perspective to understand the redundancy of SNNs by analyzing the \rone{relationship between \emph{spike firing} and \emph{spatio-temporal dynamics} of spiking neurons}. This analysis could be extended by asking three key questions. (\romannumeral1) \rone{\textbf{Which} spikes are redundant?} (\romannumeral2) \rone{\textbf{Why} is there redundancy in SNNs?} (\romannumeral3) \textbf{How} to efficiently drop the redundant spikes?

\rone{To perfectly demonstrate redundancy in SNNs, we select event-based vision tasks to observe spike responses. Event-based cameras, such as the Dynamic Vision Sensor (DVS) \cite{lichtsteiner2008128}, are a novel class of bio-inspired vision sensors that only encode the vision scene's brightness change information into a stream of events (spike with address information) for each pixel. As shown in Figure~\ref{fig:overview_of_paper}\textbf{b}, the red and green dots represent pixels that increase and decrease in brightness, respectively, and the gray areas without events indicate no change in brightness. However, although the information given in the input is human gait without background, some spike features extracted by the vanilla SNN focus on background information. As depicted in Figure~\ref{fig:overview_of_paper}\textbf{c}, the spiking neurons in the noise feature map fire a large number of spikes in the background region, which are redundant.}

\rone{Unfortunately, noise features exist widely in both temporal and spatial dimensions, but exhibit some interesting regularities.} We argue that the underlying reason for this phenomenon is due to a fundamental assumption of SNNs, known as \emph{spatio-temporal invariance}\cite{huang2021tada}, which enables sharing weights for every location across all timesteps. This assumption improves the parameter utilization efficiency while boosting the redundancy of SNNs. \rone{Specifically, by controlling the input time window of event streams, we can clearly observe the temporal and spatial changes of the spike features extracted by the SNN (see Figure~\ref{fig:spatio_temporal_invariance}). In the spatial dimension, there are many similar noise features, which can be referred to as ghost features \cite{han2020ghostnet,han2022ghostnets}. In the temporal dimension, although the information extracted by SNN changes at different timesteps, the spatial position of the noise spike feature is almost the same.} 


\rone{Recently, several works \cite{guo2022recdis,guo2022reducing,guo2022imloss} have investigated the information loss caused by SNNs when quantizing continuous membrane potential values into discrete spikes. Inspired by these works, we transformed our problem ``the relationship between spike firing and spatio-temporal dynamics of spiking neurons" to investigate the relationship between membrane potential distribution and redundant spikes.} Motivated by the observations that redundancy is highly correlated with spike feature patterns and neuron location, we present the Advanced Spatial Attention (ASA) module for SNNs, which can convert noise features into normal or null (without spike firing) features by shifting the membrane potential distribution. We conduct extensive experiments using a variety of network structures to verify the superiority of our method on five event-based datasets. \rone{Experimental results show that the ASA module can help SNN reduce spikes and improve task performance concurrently.} For instance, on the DVS128 Gait-day dataset\cite{wang_gait_cvpr_2019}, at the cost of negligible additional parameters and computations, the proposed ASA module decreases the baseline model's spike counts by 78.9\% and increases accuracy by +5.0\%. We summarize our contributions as follows:
\begin{itemize}
\item [1)] 
\rone{We provide the first systematic and in-depth analysis of the inherent redundancy in SNNs by asking and answering three key questions, which are crucial to the high energy efficiency of spike-based neuromorphic computing but have long been neglected.}
\item [2)] 
\rone{For the first time, we relate the redundancy of SNNs to the distribution of membrane potential, and design a simple yet efficient advanced spatial attention to help SNN optimize the membrane potential distribution and thus reduce redundancy.}
\item [3)]
\rone{Extensive experimental results show that the proposed ASA module can improve SNNs' performance and significantly drop noise spikes concurrently. This inspires us that two of the most important nature of spike-based neuromorphic computing, bio-inspired spatio-temporal dynamics and event-driven sparse computing, can be naturally incorporated to achieve better performance with lower energy consumption.} 
\end{itemize}

\section{Related work}
\textbf{Event-based vision and spike-based neuromorphic computing.} Due to the unique advantages of high temporal resolution, high dynamic range, etc., DVS has broad application prospects in special visual scenarios, such as high-speed object tracking \cite{zhao_2022_hybrid_framework}, low-latency interaction \cite{amir_Gesture_dataset_2017}, etc. Event-based vision is one of the typical advantage application scenarios of SNNs, which can process event streams event-by-event to achieve minimum latency \cite{Gallego_2020_DVS_Survey}, and can be smoothly deployed on neuromorphic chips to realize ultra-low energy cost by spike-based event-driven sparse computing \cite{schuman2022opportunities,rao2022long,Nature_2,eshraghian_2021_training_lesson}. As an example, a recent edge computing device called Speck\footnote {https://www.synsense-neuromorphic.com/products/speck/} integrates an SNN-enabled asynchronous neuromorphic chip with a DVS camera \cite{Gallego_2020_DVS_Survey}. Its peak power is mW level, and latency is ms level. In this work, we investigate the SNNs' inherent redundancy by using a variety of event-based datasets to further explore their enticing potential for accuracy and energy efficiency.

\textbf{Attention in SNNs.} Attention methods were included in deep learning with tremendous success and were motivated by the fact that humans can focus on salient vision information in complicated scenes easily and efficiently. A popular research direction is to present attention as an auxiliary module to boost the representation capacity of ANNs \cite{SE_PAMI,CBAM,SimAM,li2022ham,guo_2022_attention_CV_survey}. In line with this idea, Yao \etal \cite{yao_2021_TASNN} first suggested using an extra plug-and-play temporal-wise attention module for SNNs to bypass a few unnecessary input timesteps. Subsequently, a number of works were given to utilize multi-dimensional attention modules for SNNs, including temporal-wise, spatial-wise , or channel-wise simultaneously \cite{liu2022att_snn_1,zhu2022tcja,yu2022stsc,yao2022attention,YAO2023410}, where Yao \etal \cite{yao2022attention} highlighted that attention could aid SNNs in reducing spike firing while enhancing accuracy. However, to produce attention scores and refine membrane potentials, multi-dimensional attention inevitably adds a lot of extra computational burden to SNNs. In this work, we exclusively employ spatial attention, which is inspired by the investigation of the redundancy of SNNs.


\textbf{Membrane Potential Distribution in SNNs.} Rectifying MPD is crucial for SNN training because SNNs are more vulnerable to gradient vanishing or explosion since spikes are discontinuous and non-differentiable. Around this point, researchers have made many advances in SNN training, such as normalization techniques \cite{zheng_Going_Deeper_SNN_2021,wu2019direct}, shortcut design \cite{Hu_2021_MS,fang_deep_snn_2021}, extension with more learnable parameter \cite{Fang_2021_ICCV,shaban2021adaptive}, distribution loss design \cite{guo2022recdis,guo2022reducing}, etc. We here, in contrast to prior publications, concentrate on the connection between MPD and redundancy, a topic that is typically disregarded in the SNN community. 

\section{SNN Redundancy Analysis}\label{sec:MPD_analysis}

\subsection{SNN Fundamentals}
The basic computational unit of a SNN is the spiking neuron, which is the abstract modeling of the dynamics mechanism of biological neuron \cite{herz2006modeling}. The Leaky Integrate-and-Fire (LIF) model \cite{Maass_1997_LIF} is one of the most commonly used spiking neuron models since it establishes a good balance between the simplified mathematical form and the complex dynamics of biological neurons. We describe the LIF-SNN layer in its iterative representation version form \cite{Wu_STBP_2018}. First, the LIF layer will perform the following integration operations,
\begin{equation}
    \boldsymbol{U}^{t, n}=\boldsymbol{H}^{t-1, n}+\boldsymbol{X}^{t, n}, \\
    \label{eq:integrate}
\end{equation}
where $n\in\left\{1,\cdots,N\right\}$ and $t\in\left\{1,\cdots,T\right\}$ denote the layer and timestep, $\boldsymbol{U}^{t, n}$ means the membrane potential which is produced by coupling the spatial feature $\boldsymbol{X}^{t, n}$ and the temporal information $\boldsymbol{H}^{t-1, n}$, and $\boldsymbol{X}^{t, n}$ can be done by convolution operations,
\begin{equation}
    \boldsymbol{X}^{t, n} = \operatorname{BN}\left(\operatorname{Conv}\left(\boldsymbol{W}^{n}, \boldsymbol{S}^{t, n-1}\right)\right), \\
    \label{eq:Conv-based}
\end{equation}
where $\operatorname{BN}(\cdot)$ and $\operatorname{Conv}(\cdot)$ mean the batch normalization\cite{ioffe_batchNorm_2015} and convolution operation respectively, $\boldsymbol{W}^{n}$ is the weight matrix, $\boldsymbol{S}^{t, n-1}(n \neq 1)$ is a spike tensor from the last layer that only contain 0 and 1, and $\boldsymbol{X}^{t, n} \in\mathbb{R}^{c_{n} \times h_{n} \times w_{n}}$. Then, the fire and leaky mechanism inside the spiking neurons are respectively executed as
\begin{equation}
    \boldsymbol{S}^{t, n}=\operatorname{Hea}\left(\boldsymbol{U}^{t, n}-V_{th}\right), \\
    \label{eq:fire}
\end{equation}
and 
\begin{equation}
    \boldsymbol{H}^{t, n}=V_{reset}\boldsymbol{S}^{t, n} + \left(\beta \boldsymbol{U}^{t, n}\right) \otimes \left(\mathbf{1}-\boldsymbol{S}^{t, n}\right), \\
    \label{eq:leak}
\end{equation}
where $V_{th}$ is the threshold to determine whether the output spike tensor $\boldsymbol{S}^{t, n}$ should be spike or stay as zero, $\operatorname{Hea}(\cdot)$ is a Heaviside step function that satisfies $\operatorname{Hea}\left(x\right)=1$ when $x\geq0$, otherwise $\operatorname{Hea}\left(x\right)=0$, $V_{reset}$ denotes the reset potential which is set after activating the output spike, and $\beta = e^{-\frac{d t}{\tau}} < 1$ reflects the decay factor, $\tau$ is the membrane time constant, and $\otimes$ denotes the element-wise multiplication. When the entries in $\boldsymbol{U}^{t, n}$ are greater than the threshold $V_{th}$, the spatial output of spike sequence $\boldsymbol{S}^{t, n}$ will be activated (Eq.~\ref{eq:fire}). Meanwhile, the entries in $\boldsymbol{U}^{t, n}$ will be reset to $V_{reset}$, the temporal output $\boldsymbol{H}^{t, n}$ should be decided by $\boldsymbol{X}^{t, n}$ since $\mathbf{1}-\boldsymbol{S}^{t, n}$ must be 0. Otherwise, the decay of $\boldsymbol{U}^{t, n}$ will be used to transmit the $\boldsymbol{H}^{t, n}$, since the $\boldsymbol{S}^{t, n}$ is 0, which means there is no activated spike output (Eq.~\ref{eq:leak}). 

\subsection{Redundancy Analysis}\label{subsec:redundancy_analysis}
We first define various terms to appropriately represent redundancy in SNNs, as below.

\textbf{Definition 1.} \emph{Spike Firing Rate (SFR)}: We input all the samples on the test set into the network and count the spike distribution. We define a Neuron's SFR (N-SFR) at the $t$-th timestep as the ratio of the number of samples generating spikes on this neuron to the number of all tested samples. Similarly, at the $t$-th timestep, we define the SFR of a Channel (C-SFR) or this Timestep (T-SFR) as the average of the SFR values of all neurons in this channel or the entire network at this timestep. We define the Network Average SFR (NASFR) as the average of T-SFR over all timesteps $T$. 

\textbf{Definition 2.} \emph{Spike features.} We input all the samples on the test set into the network and define the average output of a channel at the $t$-th timestep as a spike feature, with each pixel's value being N-SFR. 

\textbf{Definition 3.} \emph{Ghost features.} There are numerous feature map pairs that resemble one another like ghosts \cite{han2020ghostnet,han2022ghostnets}. We call these feature maps ghost features.

\textbf{Definition 4.} \emph{Spike patterns.} Spike features display a variety of patterns, and various patterns extract different types of information. \rone{We empirically refer to the features that focus on background information as the noise pattern since there is no background information in the input}, and collectively refer to other features as the normal pattern.

\rone{Based on these definitions, we investigate the redundancy of SNNs in four granularities: spatial, temporal, channel, and neuron.}

\textbf{Observation 1.} \emph{In the spatial granularity, there are lots of ghost features in the spike response.}

\begin{figure}[!tbp]
\centering
\includegraphics[scale=0.4]{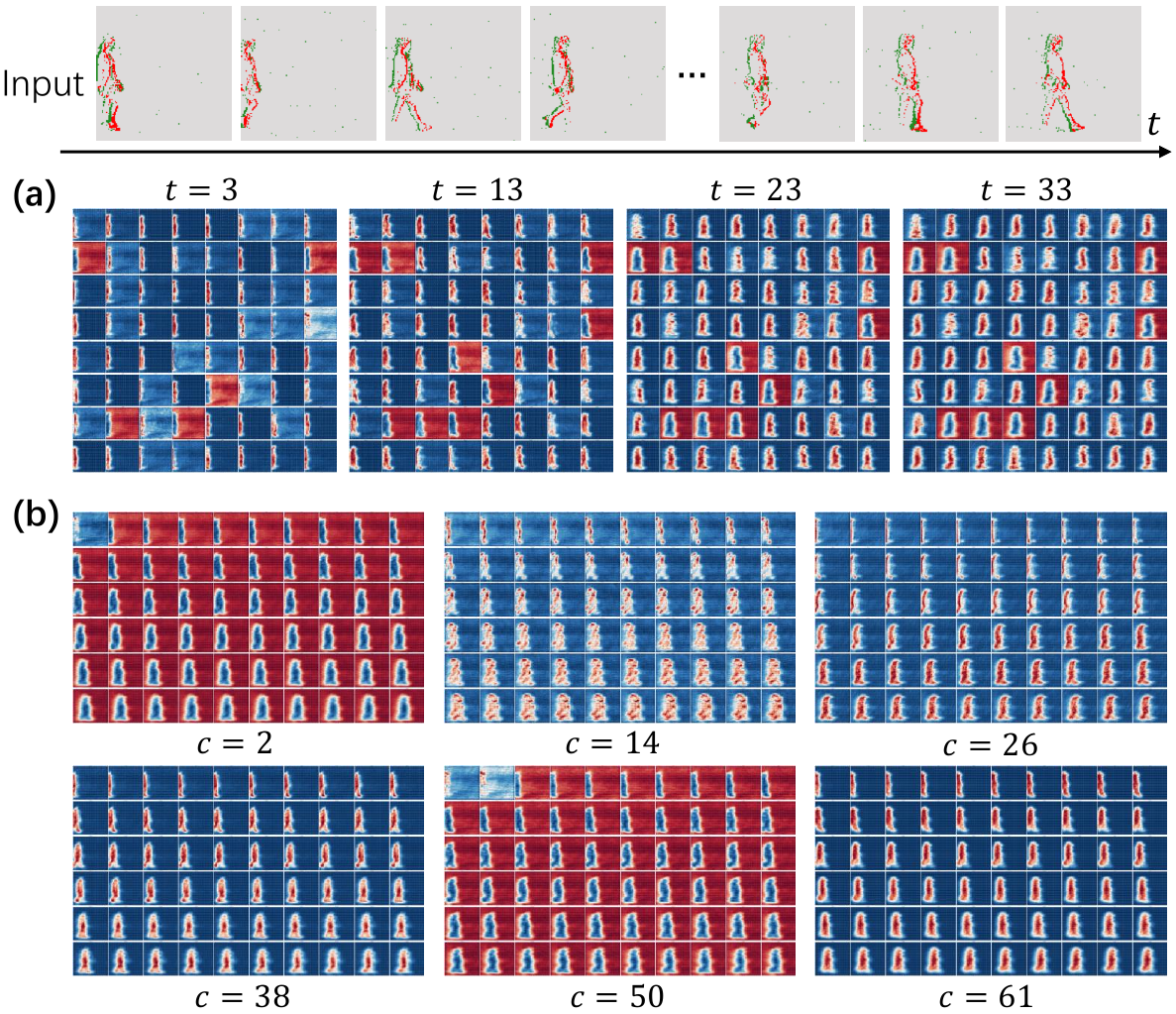}
\caption{Inherent redundancy in SNNs exists in both spatial and temporal granularities, originating from network \emph{over-parameterization} and \emph{spatio-temporal invariance}, respectively. (a) spike features (averaging the spike tensors $\boldsymbol{S}^{t, n}$ over all samples) of different channels at the same timestep. Each pixel indicates the firing rate of a spiking neuron. The bluer the pixel, the closer the firing rate is to 0; the redder the pixel, the closer the firing rate is to 1. In the noise features, the background region is big and nearly totally red, indicating that many spikes are produced. (b) spike features of the same channel at different timesteps.}
\label{fig:spatio_temporal_invariance}
\end{figure}

\rone{Redundancy is inevitable in over-parameterized neural networks. For instance, from the perspective of feature maps, there are many ghost features in Conv-based ANNs (CNNs) \cite{han2020ghostnet,zhou_2016_CAM}. The same is true for SNNs}, as demonstrated in Figure~\ref{fig:spatio_temporal_invariance}\textbf{a}. Plotting all spike features at the same timestep, we see that certain features concentrate on background information with a huge region of red, while others concentrate on gait information with a large area of blue, and ghost features can be seen in both patterns.

\textbf{Observation 2.} \emph{In the temporal granularity, T-SFR at different timesteps does not change much.}

Given that each timestep shares the weight of 2D convolution for spatial modeling, the level of redundancy has increased significantly for SNNs that do temporal modeling. To show this, we give spike features of the same channel at various timesteps in Figure~\ref{fig:spatio_temporal_invariance}\textbf{b}. We see that for a fixed channel, the features derived at various timesteps differ, i.e., the human gait shifts progressively to the right as the timestep increases. Interestingly, the same channel's spike features—almost all of which is background information or all of which is information on human gait—are essentially the same at different timesteps. This demonstrates that spatio-temporal invariance will result in similar spike features at different timesteps. It also implies that redundancy in SNNs is linearly connected to timesteps. 

\rone{\textbf{Observation 3.} \emph{In the channel granularity, the C-SFR is closely related to the spike patterns learned by this channel. In the neuron granularity, the N-SFR is tightly linked to the location of neurons.}}

We zoom in to highlight two typical spike features that pertain to various patterns in Figure~\ref{fig:overview_of_paper}\textbf{c}. We see that the two features have substantially different C-SFRs. The spike feature of the noise pattern fires many spikes while focusing on trivial background information. By contrast, the spike feature of the normal pattern with lower C-SFR focuses on salient gait information in a condensed region. Furthermore, the N-SFRs of neurons in the background region of normal features are almost zero, but the N-SFRs of neurons in the same region in noise features are very high.

\textbf{Definition 5.} \emph{Membrane Potential Distribution (MPD).} We input all the samples on the test set into the network. In the $c$-th channel of the $n$-th layer, we count the membrane potential values of all neurons at the $t$-th timestep. We can represent the membrane potential distribution of the channel by a 2D histogram, where the horizontal axis is the value of the membrane potential, and the vertical axis is the number of neurons located in a certain window. 


\textbf{Observation 4.} \emph{Membrane potential distributions are highly correlated with spike patterns.}

\rone{Note, to facilitate the analysis of the effect of the proposed method on the MPD and the spike feature, we discuss them in detail later in Section~\ref{subsec:MPD_analysis}.}

\section{Methodology}\label{sec:method}
\textbf{Motivation.} We can infer the following three empirical conclusions from : (1) Features can basically be separated into noise, and normal patterns; (2) The quality of the feature is determined by the MPD of the channel; (3) The firing of spiking neurons is related to their location. Based on these observations, we concentrate our optimization on the MPD of each channel, i.e., performing spatial attention. Meanwhile, considering that all features have two patterns, we exploit independent spatial attention to optimize them separately, called advanced spatial attention.

\begin{figure}
\centering
\begin{subfigure}{0.8\linewidth}
{\includegraphics[scale=0.48]{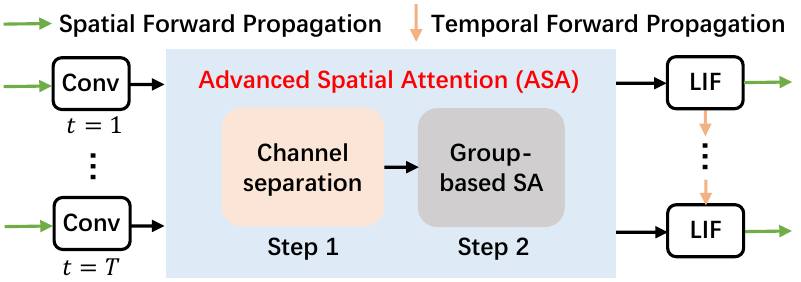}}
\caption{Overview of ASA-SNN.}
\label{fig:network_structure_0}
\end{subfigure}

  \begin{subfigure}{0.8\linewidth}
    {\includegraphics[scale=0.45]{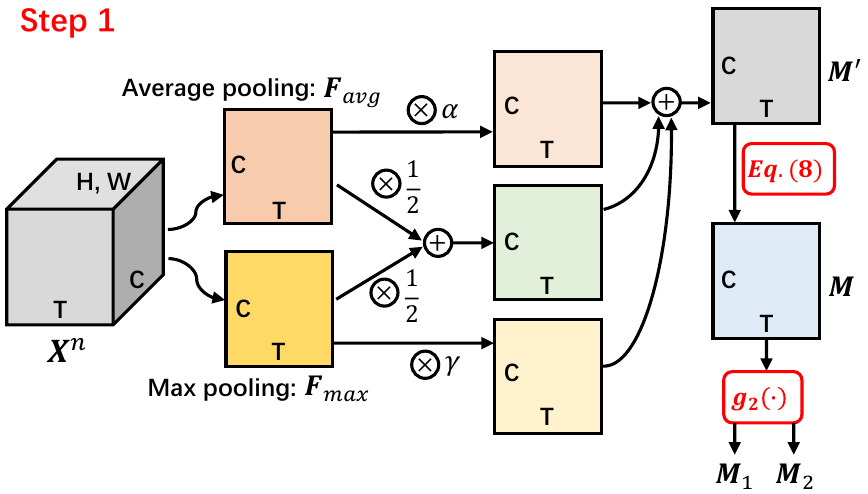}}
    \caption{Channel separation technique (Eq.~\ref{eq:mask}).}
    \label{fig:network_structure_1}
  \end{subfigure}

  \begin{subfigure}{0.87\linewidth}
    {\includegraphics[scale=0.45]{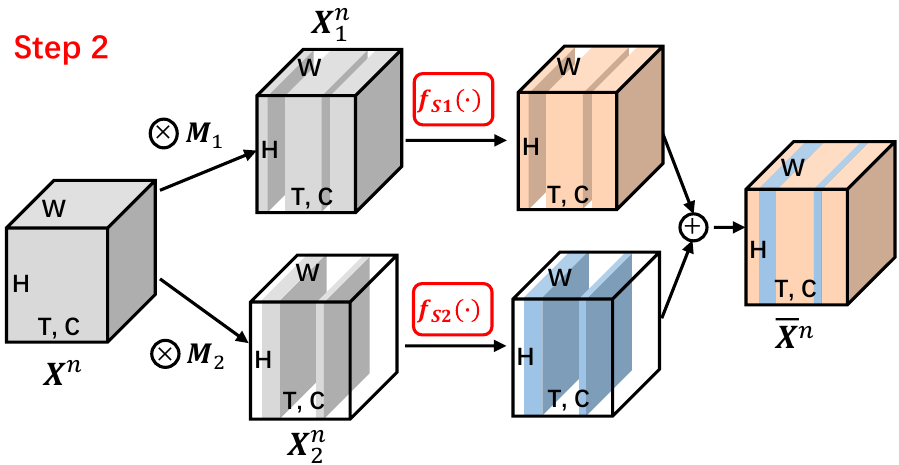}}
    \caption{Group-based SA (Eq.~\ref{eq:SA}).}
    \label{fig:network_structure_2}
  \end{subfigure}
  \caption{Details of ASA-SNN. The ASA module is divided into two steps: (b) Channel separation and (c) Group-based SA, and consists of four functions, $g_1(\cdot)$, $g_2(\cdot)$, $f_{S1}(\cdot)$, and $f_{S2}(\cdot)$.}
  \label{fig:network_structure}
\end{figure}

\textbf{Method.} As shown in Figure~\ref{fig:network_structure}\textbf{a}, we implement our ASA module in two steps. We first exploit a channel separation technique to separate all features into two complementary groups based on their importance. Then individual SA sub-modules are performed on the two groups of features. Suppose $\boldsymbol{X}^{n} = \left[\cdots, \boldsymbol{X}^{t,n}, \cdots \right] \in \mathbb{R}^{T \times c_{n} \times h_{n} \times w_{n}}$ is an intermediate feature map as input tensor, this two-step process can be summarized by the following equations:
\begin{equation}
    \boldsymbol{M}_{1}, \boldsymbol{M}_{2} = g_{2}(g_{1}(\boldsymbol{X}^{n})), \\
    \label{eq:mask}
\end{equation}
\begin{equation}
    \overline{\boldsymbol{X}}^{n} = f_{S1}(\boldsymbol{X}^{n}\otimes \boldsymbol{M}_{1}) \oplus f_{S2}(\boldsymbol{X}^{n}\otimes \boldsymbol{M}_{2}), \\
    \label{eq:SA}
\end{equation}
where $\boldsymbol{M}_{1}, \boldsymbol{M}_{2} \in \mathbb{R}^{T \times c_{n} \times 1 \times 1}$ are the complementary mask (separation) maps that contain only 0 and 1 elements, $g_1(\cdot)$ is a function that assesses the channel's importance, $g_2(\cdot)$ is the separation policy function used to generate mask maps for feature grouping, $f_{S1}(\cdot)$ and $f_{S2}(\cdot)$ are SA functions with the same expression, $\overline{\boldsymbol{X}}^{n}$ is the output feature tensor which has the same size as $\boldsymbol{X}^{n}$. During multiplication, the mask score are broadcast (copied) along the temporal and channel dimensions accordingly. Finally, compared with $\boldsymbol{U}^{t, n}$ of vanilla SNN in Eq.~\ref{eq:integrate}, the \emph{new membrane potential} behaviors of ASA-SNN layer follow
\begin{equation}
    \boldsymbol{U}^{t, n}=\boldsymbol{H}^{t-1, n}+\overline{\boldsymbol{X}}^{t, n}. \\
    \label{eq:ASA-SNN}
\end{equation}

Empirically, the design of $g_1(\cdot)$ is critical to the task accuracy, as well as the number of additional parameters and computations. The classic channel attention models in CNNs \cite{SE_PAMI,CBAM,Wang_2020_ECA,SimAM,li2022ham} generally judge the importance of the channel by fusing the global degree information (max pooling) and local significance information (average pooling) of the features. Inspired by these works, here we design two schemes for $g_1(\cdot)$, one that is learnable (\emph{ASA-1}) and the other that directly judges importance based on pooled information (\emph{ASA-2}).

As shown in Figure~\ref{fig:network_structure}\textbf{b}, temporal-channel features are aggregated by using both average-pooling and max-pooling operations, which infer two different tensors $\boldsymbol{F}_{avg}, \boldsymbol{F}_{max} \in \mathbb{R}^{T \times c_{n} \times 1 \times 1}$. 

In ASA-1, we get the importance map $\boldsymbol{\boldsymbol{M}}$ by
\begin{equation}
    \boldsymbol{M}’ = \frac{1}{2} \otimes (\boldsymbol{F}_{avg} + \boldsymbol{F}_{max}) + \alpha \otimes \boldsymbol{F}_{avg} + \gamma \otimes \boldsymbol{F}_{max},
\end{equation}
\begin{equation}
    \boldsymbol{M} = \sigma\left(\boldsymbol{W}_{2}^{n}(\operatorname{ReLU}(\boldsymbol{W}_{1}^{n}(\boldsymbol{M}’))))\right.,
    \label{eq:SE}
\end{equation}
where $\alpha$ and $\gamma$ are trainable parameters which are initialised with 0.5, $\sigma$ means the sigmoid function, $\boldsymbol{W}_{1}^{n}\in \mathbb{R}^{\frac{T}{r} \times T}$ and $\boldsymbol{W}_{2}^{n}\in \mathbb{R}^{T \times \frac{T}{r}}$ \rone{are trainable parameters independent at each layer}, and $r$ represents the dimension reduction factor. Note, $\boldsymbol{M}’, \boldsymbol{M} \in \mathbb{R}^{T \times c_{n} \times 1 \times 1}$, we share $\boldsymbol{W}_{1}^{n}$ and $\boldsymbol{W}_{2}^{n}$ on the channel dimension. 

In ASA-2, we set $\boldsymbol{M} = \boldsymbol{M}’$ directly. Then, we get $\boldsymbol{M}_{1}$ and $\boldsymbol{M}_{2}$, denoting the important and sub-important channel indexes respectively, by combining the $y$-th largest values of two dimensions in $\boldsymbol{M}$. Specifically, the pseudo-code of $g_2(\cdot)$ is represented by

\begin{minted}
[
frame=lines,
framesep=2mm,
baselinestretch=1.2,
fontsize=\footnotesize,
]
{python}
# X: input feature [N, T, C, H, W]
# k: 0.5 * C
def select_max(X, dim="C", k=k):
    mask = zeros_like(X)
    mask[topk(X, dim=dim, k=k)] = 1
    return mask

def mask(X, k):
    mask_c = select_max(X, dim="C", k=k)
    mask_t = select_max(X, dim="T", k=k)
    mask = (mask_c + mask_t) / 2
    return where(mask == 0.5, 1, 0)
\end{minted}

\begin{figure}[!htb]
\setlength{\belowcaptionskip}{-0.3cm}
\centering
\includegraphics[scale=0.6]{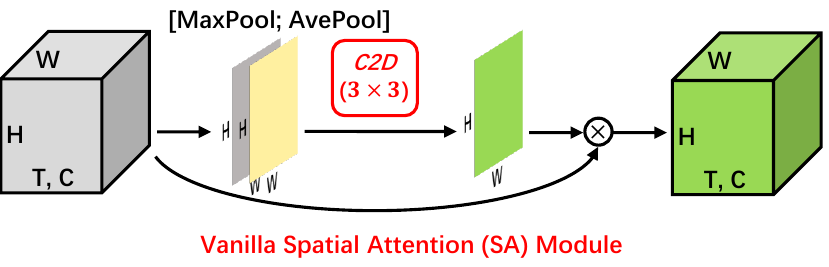}
\vspace{-2mm}
\caption{Diagram of spatial attention module. As illustrated, the spatial attention exploits two outputs that are pooled along the temporal-channel axis and forward them to a $3\times3$ convolution layer.}
\label{fig:vanilla_SA}
\end{figure}

After obtaining $\boldsymbol{X}_1^{n} = \boldsymbol{X}^{n}\otimes \boldsymbol{M}_{1}$ and $\boldsymbol{X}_2^{n} = \boldsymbol{X}^{n}\otimes \boldsymbol{M}_{2}$, we perform individual SA module to optimize them. As shown in Fig.~\ref{fig:vanilla_SA}, the SA is follow \cite{CBAM}:
\begin{equation}
    f_{S}(\cdot) = \sigma(f^{3\times3}([MaxPool(\cdot);AvgPool(\cdot)])), \\
    \label{eq:ASA}
\end{equation}
where $\operatorname{AvgPool}(\cdot), \operatorname{MaxPool}(\cdot) \in \mathbb{R}^{1 \times 1 \times h_{n} \times w_{n}}$, $f^{3 \times 3}$ represents a convolution operation with the filter size of $3 \times 3$, $f_{S}(\cdot) \in \mathbb{R}^{1 \times 1 \times h_{n} \times w_{n}}$ is the 2-D spatial attention scores, and we set $f_{S1}(\cdot) = f_{S2}(\cdot) = f_{S}(\cdot)$.

\section{Experiments}
For an event stream, we exploit the frame-based representation \cite{yao_2021_TASNN,Fang_2021_ICCV} as the preprocessing method to convert it into an event frame sequence. Suppose the interval between two frames (i.e., temporal resolution) is $dt$ and there are $T$ frames (i.e., timesteps), the length of the input event stream is $t_{lat} = dt \times T$ millisecond. After processing these divided frames through SNN, a prediction can then be retrieved. 

\subsection{Experimental Setup}\label{subsec:experiment_setup}
We evaluate our method on five datasets, all generated by recording actions in real scenes. DVS128 Gesture \cite{amir_Gesture_dataset_2017}, DVS128 Gait-Day \cite{wang_gait_cvpr_2019}, and DVS128 Gait-Night \cite{wang_Gait_PAMI_2022} were captured by a 128x128 pixel DVS128 camera. As their names imply, Gesture comprises hand gestures, while Gait-day and Gait-night include human gaits in daylight and at night, respectively. DailyAction-DVS \cite{liu2021event} and HAR-DVS \cite{wang2022hardvs} are acquired by a DAVIS346 camera with a spatial resolution of 346x260, of which HAR-DVS has 300 classes and 107,646 samples and is currently the \emph{largest} event-based human activity recognition (HAR) dataset. The raw HAR-DVS exceeds 4TB. The authors convert each event stream into frames and randomly sample 8 frames to form a new HAR-DVS for ease of processing. 

We execute the baseline for each group of ablation trials, then plug the proposed ASA to run the model again (Table~\ref{tab:main_result}). Each group of trials for vanilla and ASA- SNNs employed the same hyper-parameters, training methods, and other training conditions\footnote{Details of datasets and training are given in the Supplementary.}. In all experiments, we exploit a total of three baselines with different structures. We carefully selected baselines for various datasets to examine the relationship between the spike firing, the dataset, and the network structure. One is the shallow three-layer Conv-based LIF-SNN presented in \cite{yao_2021_TASNN,yao2022attention}. The other is a deeper five-layer Conv-based LIF-SNN, following \cite{Fang_2021_ICCV}. Finally, the Res-SNN-18 \cite{fang_deep_snn_2021} in the SpikingJelly framework\footnote{https://github.com/fangwei123456/spikingjelly} is used to verify the large datasets.

\begin{table}[t]
\setlength{\tabcolsep}{0.1mm}
    \centering
    \renewcommand{\arraystretch}{1.5}
    \resizebox{\linewidth}{!}{
    \setlength{\tabcolsep}{1.3mm}{
    \begin{tabular}{cccc}
    \hline
    Dataset & Model & Acc.(\%) & NASFR \\ \hline
    \multirow{4}{*}{Gesture} & LIF-SNN \cite{yao_2021_TASNN} & 91.3 & 0.176 \\
            & \textbf{+ ASA (Ours)} & \textbf{95.2(+3.9)} & 0.038(\textbf{-78.4\%}) \\ \cline{2-4}
            & LIF-SNN \cite{Fang_2021_ICCV} & 95.5 & 0.023 \\
            & \textbf{+ ASA (Ours)} & \textbf{97.7(+2.2)} & 0.018(\textbf{-21.7\%}) \\ \hline
    \multirow{2}{*}{Gait-day} & LIF-SNN \cite{yao_2021_TASNN} & 88.6 & 0.214 \\
            & \textbf{+ ASA (Ours)} & \textbf{93.6(+5.0)} & 0.045(\textbf{-78.9\%})\\ \hline
    \multirow{2}{*}{Gait-night} & LIF-SNN \cite{yao_2021_TASNN} & 96.4 & 0.197 \\
    & \textbf{+ ASA (Ours)} & \textbf{98.6(+2.2)} & 0.126(\textbf{-36.0\%}) \\ \hline
    \multirow{2}{*}{\makecell[c]{{DailyAction} \\ {-DVS}}} & LIF-SNN \cite{Fang_2021_ICCV} & 92.5 & 0.017 \\
    & \textbf{+ ASA (Ours)} & \textbf{94.6(+2.1)} & 0.013(\textbf{-23.5\%}) \\ \hline
    \multirow{2}{*}{HAR-DVS} & Res-SNN-18 \cite{fang_deep_snn_2021} & 45.5 & 0.206 \\
    
           & \textbf{+ ASA (Ours)} & \textbf{47.1(+1.6)} & 0.183(\textbf{-11.2\%}) \\ \hline    
    \end{tabular}
    }
    }
    \vspace{-2mm}
    \caption{Main results of vanilla vs. ASA- SNNs (ASA-1). Except for HAR-DVS, reported accuracies are average of five replicates.}
    \label{tab:main_result}
\end{table}

\subsection{Ablation Study for ASA Module}\label{subsec:main_result}
In terms of accuracy and NASFR, We present the main results in Table~\ref{tab:main_result}. ASA-SNN achieves higher task accuracy with lower spike firing in all ablation studies. The performance and energy gains are more noticeable, particularly when the network structure is small. For example, in the Gait-day, plugging the ASA module into a three-layer SNN \cite{yao_2021_TASNN} can reduce the spike counts by 78.9\% and improve the performance by +5.0 percent. This is crucial for the deployment of SNN algorithms on neuromorphic chips, which usually have strict memory limitations \cite{davies2018loihi,Nature_1,Nature_2}. The ASA module also performs well on the deep Res-SNN. For instance, on HAR-DVS, the ASA-SNN outperforms the original SNN +1.7 percent while firing fewer spikes. \rone{In addition, we provide more ablation studies on the ASA module in the Supplementary.} 

\rone{Although it is beyond the scope of this work, by observing results in Table~\ref{tab:main_result}, we raise another complex and important question: ``\emph{What factors affect the redundancy of SNN?}" Intuitively, we could exploit NASFR as a redundancy indicator for SNNs. We argue that the NASFR of SNNs depends on various factors, the core of which includes dataset size, network size, spiking neuron types, etc. For instance, on Gesture, the NASFRs in three-layer \cite{yao_2021_TASNN} and five-layer vanilla SNN \cite{Fang_2021_ICCV} are 0.176 and 0.023, respectively. Empirically, vanilla SNN's NASFR also affects the function of the ASA module, where SNNs with more redundancy may be easier to reduce spikes. We hope that these observations will inspire more theoretical and optimization work on redundancy.}

\subsection{Comparison with the State-of-the-Art}
In Table~\ref{Table:Compare_with_prior_work}, we make a comprehensive comparison with prior works in terms of input temporal window and accuracy. Since some datasets were created recently, there is a lack of benchmarks in the field of SNNs. In this paper, we benchmark these datasets using models from the open-source framework SpikingJelly and fill in the corresponding accuracies in Table~\ref{tab:main_result} and Table~\ref{Table:Compare_with_prior_work}. We can see that on four small datasets, ASA-SNN can produce SOTA or comparable performance. Compared to GCN methods \cite{wang_gait_cvpr_2019,wang_Gait_PAMI_2022} with full input, we observe that SNNs can always achieve higher performance with less input (i.e., smaller $dt \times T$). \rone{Moreover, on the largest HAR-DVS dataset, our Top-1 accuracy is 47.1\% based on Res-SNN-18, which is comparable to the ANN-based benchmark results from 46.9\% to 51.2\%. This is a reasonable result since SNNs employ binary spikes, generally gaining higher energy efficiency at the expense of accuracy.}

\begin{table}[t]
\setlength{\tabcolsep}{0.1mm}
\renewcommand{\arraystretch}{1.5}
\centering
\resizebox{\linewidth}{!}{
\setlength{\tabcolsep}{1.3mm}{
\begin{tabular}{cccc}
\hline
Dataset & Methods & {$dt \times T$} & Acc. (\%) \\ \hline

	\multirow{5}{*}{Gesture} 
    & 12 layers CNN \cite{amir_Gesture_dataset_2017} & $1 \times 120$ & 92.6 \\ 
    & PLIF-SNN \cite{Fang_2021_ICCV} & $300 \times 20$ & 97.6 \\ 
    & Res-SNN-18 \cite{yao_2021_TASNN} & $375 \times 16$ & 97.9 \\ 
    & MA-SNN \cite{yao2022attention} & $300 \times 20$ & \textbf{98.2} \\ \cline{2-4}
    & This Work & $300 \times 20$ & 97.7 \\ \hline
 
	\multirow{5}{*}{{Gait-day}} 
    & EV-Gait GCN \cite{wang_gait_cvpr_2019}  & $4400 \times 1$  & 89.9 \\ 
    & TA-SNN \cite{yao_2021_TASNN}  & $15 \times 60$  & 88.6 \\ 
    & 3D GCN \cite{wang_Gait_PAMI_2022}  & $1500 \times 1$  & 86.0 \\ 
    & MA-SNN \cite{yao2022attention} & $15 \times 60$  & 92.3 \\ \cline{2-4}
    & This Work & $15 \times 60$ & \textbf{93.6} \\ \hline
 
	\multirow{3}{*}{{Gait-night}} 
    & TA-SNN \cite{yao_2021_TASNN} & $15 \times 60$ & 96.4 \\ 
    & 3D GCN \cite{wang_Gait_PAMI_2022} & $5500 \times 1$ & 96.0 \\ \cline{2-4}
    & This Work & $15 \times 60$ & \textbf{98.6} \\ \hline
 
        \multirow{4}{*}{\makecell[c]{{DailyAction} \\ {-DVS}}}
    & HMAX-SNN \cite{liu2020effective} & - & 76.9 \\ 
    & Motion-SNN \cite{liu2021event} & - & 90.3 \\ 
    & PLIF-SNN \cite{Fang_2021_ICCV} & $120 \times 36$ & 92.5 \\ \cline{2-4}
    & This Work & $120 \times 36$ & \textbf{94.6} \\ \hline

        \multirow{7}{*}{{HAR-DVS}} 
    & Res-CNN-18 \cite{he2016deep} & $T=8$ & 49.2 \\
    & ACTION-Net \cite{wang2021action} & $T=8$ & 46.9 \\
    & TimeSformer \cite{bertasius2021space} & $T=8$ & 50.8 \\
    & SlowFast \cite{feichtenhofer2019slowfast} & $T=8$ & 46.5 \\
    & ES-Transformer \cite{wang2022hardvs}  & $T=8$ & \textbf{51.2} \\
    & Res-SNN-18 \cite{fang_deep_snn_2021} & $T=8$ & 45.5 \\ \cline{2-4}
    & This Work & $T=8$ & 47.1 \\ \hline

	\end{tabular}
	}
	}
	\vspace{-2mm}
	\caption{The comparison between the proposed methods and existing SOTA techniques on five event-based vision datasets. Note, all the results of the ANN models in HAR-DVS in this table are taken from \cite{wang2022hardvs}. (Bold: the best) }
	\label{Table:Compare_with_prior_work}
\end{table}

\begin{table}[!t]
    \setlength{\tabcolsep}{0.1mm}
	\renewcommand{\arraystretch}{1.5}
	\centering
	\resizebox{\linewidth}{!}{
	\setlength{\tabcolsep}{1.3mm}{
	\begin{tabular}{ccccc}
	\hline
	Model & Acc. (\%) & Params ($\uparrow$) & NASFR  & $\Delta_{MAC}$ ($\uparrow$)\\
	\midrule[0.8pt]
	Vanilla SNN \cite{yao_2021_TASNN} & 88.6  & 2,323,531      & 0.214  & - \\
	+ SA \cite{yao2022attention}      & 89.5(+0.9)  & +294   & 0.091(-57.4\%)  & +14.3M \\
	+ TCSA \cite{yao2022attention}    & 92.3(+3.7)  & +24,126 & \textbf{0.045(-78.9\%)} & +27.3M \\
	+ \textbf{ASA-1 (Ours})           & \textbf{93.6(+5.0)}  & +10,914 &  \textbf{0.045(-78.9\%)} & \textbf{+2.6M} \\
	+ \textbf{ASA-2 (Ours})           & 89.6(+1.0)  & \textbf{+114}  & 0.088(-58.9\%) & \textbf{+2.6M} \\\bottomrule[1.2pt]
	\bottomrule[1.2pt]
	Vanilla SNN \cite{yao_2021_TASNN} & 91.3  & 2,323,531  & 0.176  & - \\
	+ SA \cite{yao2022attention}   & 92.6(+1.3) & +294 & 0.073(-58.5\%) & +14.3M \\
	+ TCSA \cite{yao2022attention} & 96.5(\textbf{+5.2}) & +24,126     & \textbf{0.029(-83.5\%)} & +27.3M\\
	+ \textbf{ASA-1 (Ours})     & 95.2(+3.9)    & +10,914  & 0.038(-78.4\%)  & \textbf{+2.6M}\\
	+ \textbf{ASA-2 (Ours})     & 94.4(\textbf{+3.1})    &  \textbf{+114}  & 0.050(-71.6\%) & \textbf{+2.6M} \\ 
	\bottomrule[1.2pt]
	\end{tabular}
	}
	}
	\vspace{-2mm}
	\caption{Effect of Different attention modules in three-layer SNN\cite{yao_2021_TASNN} on Gait-day (the above table) and Gesture (the below table) with $dt=15, T=60$.}
	\label{Table:compare_with_att_snn}
\end{table}


\subsection{Comparison with Other Attention SNNs}\label{subsec:compare_with_other_attention}
In this work, based on redundancy analysis, we design the ASA module, which only performs spatial attention. As mentioned, the current practice of attention mechanisms in SNNs \cite{yao2022attention,liu2022att_snn_1,zhu2022tcja,YAO2023410} is dominated by multi-dimensional composition. An easily overlooked fact is that adding attention modules inevitably introduces additional computation. These extra computations are trivial in CNNs, but require special care in SNNs, as otherwise the energy advantage of attention SNNs is lost. \rone{Specifically, the energy shift between vanilla and attention SNNs can be computed as
\begin{equation}
    \Delta_{E} = E_{MAC} \cdot \Delta_{MAC} - E_{AC} \cdot \Delta_{AC}, \\
    \label{eq:delta_E}
\end{equation}
where $E_{MAC} = 4.6pJ$ and $E_{AC} = 0.9pJ$ represent the energy cost of Multiply-and-Accumulate (MAC) and AC operation \cite{horowitz_energy_cost_2014}, $\Delta_{MAC}$ and $\Delta_{AC}$ represent the additional MAC operation and the reduced AC number caused by the attention modules, respectively (detailed energy evaluation is in the Supplementary). We need to try our best to make the benefit ($E_{AC} \cdot \Delta_{AC}$) outweigh the cost ($E_{MAC} \cdot \Delta_{MAC}$). 
}

In Table~\ref{Table:compare_with_att_snn}, we compare the number of extra parameters and computations needed for various attention modules. We see that ASA module is a cost-effective solution, less $\Delta_{MAC}$ (just 2.4M), better or comparable performance. For instance, a 78.9\% decrease in spike firing results in a 65\% reduction in energy consumption in the TCSA-SNN \cite{yao2022attention} but a 76\% reduction in our ASA-SNN. The ASA-2 design, which only adds 114 parameters to obtain a nice performance improvement on Gesture, is highlighted lastly (albeit it is not stable). 

\begin{figure}[!tbp]
\setlength{\belowcaptionskip}{-0.3cm}
\centering
\includegraphics[scale=0.36]{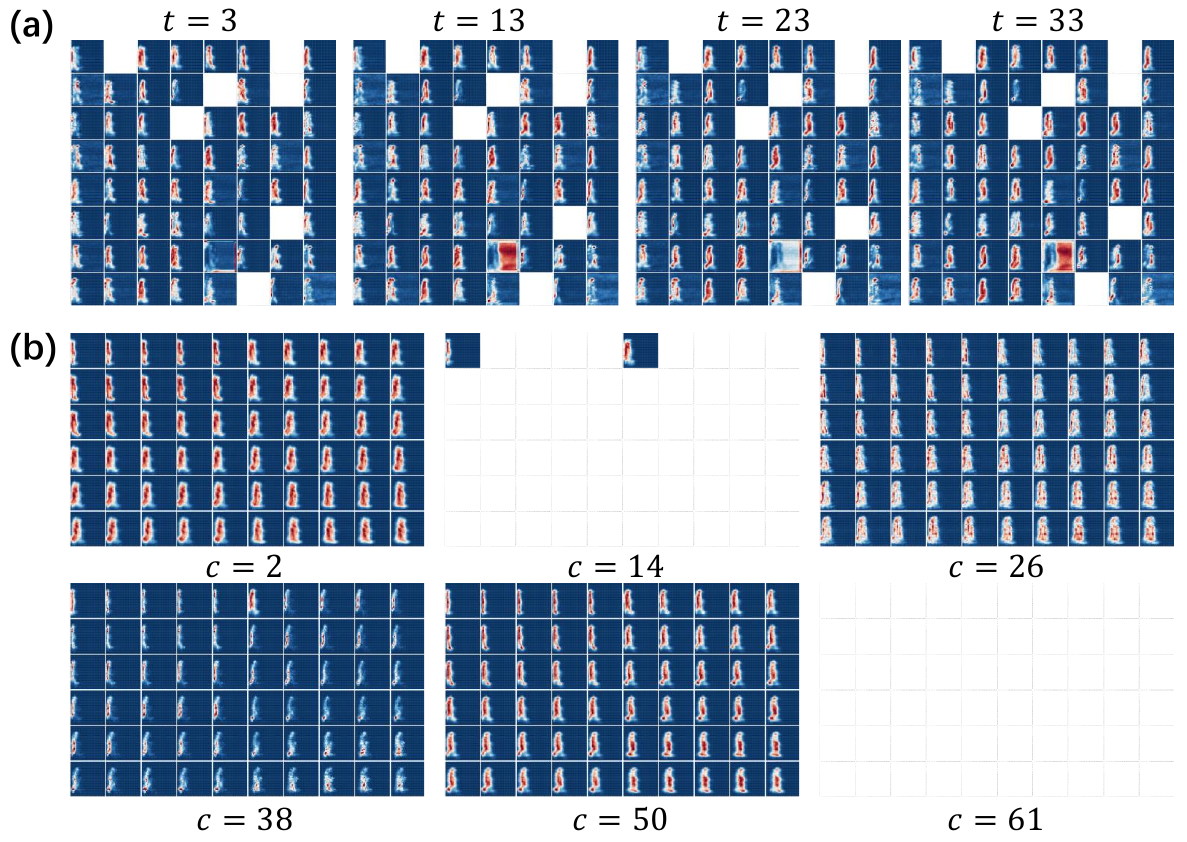}
\vspace{-2mm}
\caption{Spike features in ASA-SNN. (a) Spike features of different channels at the same timestep. (b) Spike features of the same channel at different timesteps.}
\label{fig:ASA_visualization}
\end{figure}

\begin{figure}
\centering
\begin{subfigure}{0.9\linewidth}
{\includegraphics[scale=0.4]{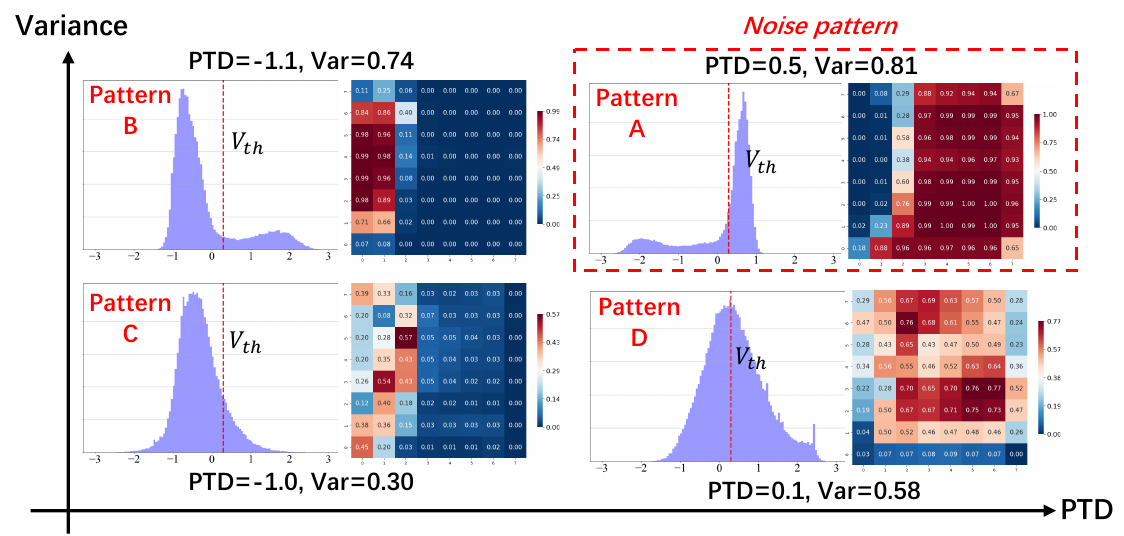}}
\caption{Spike patterns in vanilla SNNs.}
\label{fig:MPD_vanilla}
\end{subfigure}
  
\begin{subfigure}{0.9\linewidth}
{\includegraphics[scale=0.4]{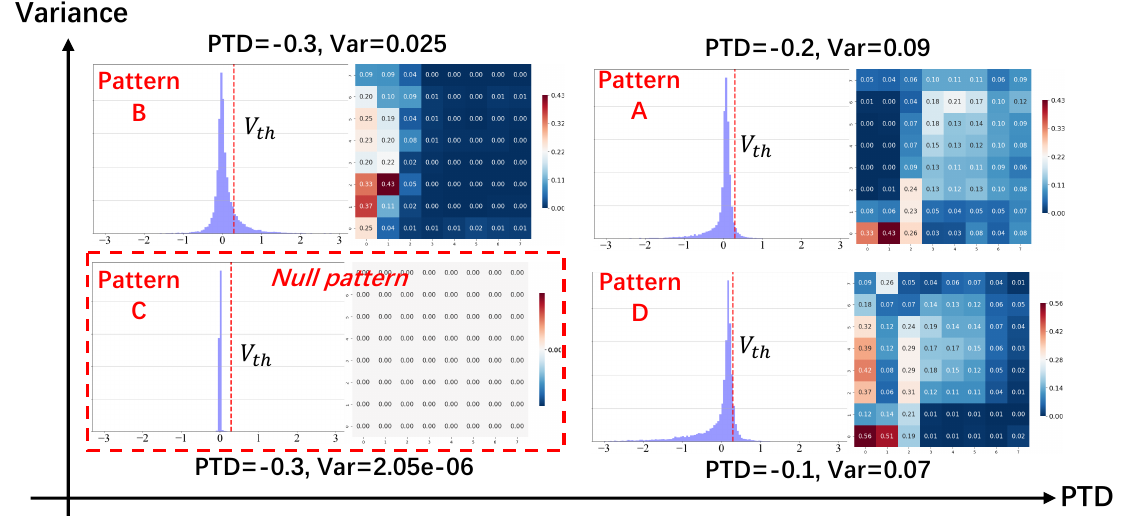}}
\caption{Spike patterns in ASA-SNNs.}
\label{fig:MPD_asa}
\end{subfigure}
\caption{When the ASA is plugged, the spike pattern shifts in membrane potential distribution and spike feature.}
 \label{fig:MPD_analysis}
\end{figure}

\rone{\subsection{Result Analysis}}\label{subsec:MPD_analysis}
\textbf{Spike patterns in ASA-SNNs.} We re-examine the spike response in ASA-SNNs as we did in Section~\ref{subsec:redundancy_analysis}. In the spatial granularity, the spike patterns in ASA-SNNs are altered. As shown in Figure~\ref{fig:ASA_visualization}\textbf{a}, there are almost no noise features, but some null features without spikes appear. In the temporal granularity, spatial-temporal invariance still holds. As depicted in Figure~\ref{fig:ASA_visualization}\textbf{b}, spike features of the same channel at different timesteps are similar.

\rone{\textbf{Membrane Potential Distribution (MPD) and spike pattern.} We already know that the redundancy in SNNs depends directly on the learned spike patterns. Therefore, we are interested in the question of ``how the spike feature changes", which can help us understand the dynamics inside the network and inspire future work. Here we analyze the relationship between the MPD and the spike feature (pattern). We define the following indicator.}

\textbf{Definition 6.} \emph{Peak-to-threshold distance (PTD).} We picked out the highest three pillars in membrane potential distribution and obtained the peak interval by averaging these pillars' membrane potential intervals. We then define peak-to-threshold distance as the difference between the center point of the peak interval and the threshold.

\textbf{Observation 5.} \emph{The PTD and Variance of membrane potential distribution of a channel can be exploited to measure the quality of the spike feature extracted by this channel to a certain extent. \rone{When the value of PTD is near 0 or greater than 0, it indicates that the membrane potential of most spiking neurons on a map is to the right of the threshold. Consequently, most neurons have a relatively high neuron spike firing rate, and intuitively, the pattern learned by the channel is background noise since the key information is usually located within a small area.} The variance measures the degree of focus. In the same or similar normal pattern of the same model, the larger the variance, the clearer the edge information of the learned feature.}

Accordingly, as shown in Figure~\ref{fig:MPD_analysis}, \rone{we show how the spike feature follows the MPD.} Specifically, in vanilla SNNs (Figure~\ref{fig:MPD_analysis}\textbf{a}), spike features and MPDs in Patterns A and B appear to be in a complementary relationship, corresponding to perfect focus on the background and object regions, respectively. If the PTD is maintained constant, as the variance gradually decreases, the information in the edge regions of the background and object begins to blur, as shown in Patterns C and D. 

Then we compare the shifts in spike features of vanilla and ASA-SNNs. Obviously, peak regions of the MPDs across all channels in ASA-SNNs are located to the left of the threshold (Figure~\ref{fig:MPD_analysis}\textbf{b}), i.e., $PTD < 0$. This indicates that one channel does not fire a lot of spikes after the ASA module has optimized the MPD. That is, the MPD is highly compact, which implies that the edge information of the spike feature is clearer. 

By combining the two indicators PTD and variance, we can quickly determine what a "good" spike feature's MPD should be. For instance, as shown in Pattern B of ASA-SNNs, both the PTD and variance values should fall within an appropriate range, neither too high nor too low.



\begin{table}[!htb]\footnotesize
    \setlength{\belowcaptionskip}{-0.3cm}
	\renewcommand{\arraystretch}{1.2}
	\centering
	\begin{tabular}{cccc}
	\toprule[1.2pt]
	Model & Gesture & Gait-day  & HAR-DVS\\
	\midrule[0.8pt]
	Vanilla SNN  & 0.158 & 0.362  & 0.584 \\ 
	ASA-SNN      & 0.024 & 0.031  & 0.339 \\ 
	\bottomrule[1.2pt]
	\end{tabular}
    \vspace{-2mm}
	\caption{Comparison of QEs on vanilla and ASA-SNNs.}
	\label{Table:QE_Value}
\end{table}

\textbf{Information loss.} As discussed in \cite{guo2022recdis,guo2022reducing}, a good MPD can reduce information loss, which arises from the quantization error (QE) introduced by converting the analog membrane potential into binary spikes. Similar to \cite{guo2022recdis,guo2022reducing}, we define the QE as the square of the difference between the membrane potential and its corresponding quantization spike value. The proposed ASA module concurrently optimizes the PTD and variance of MPDs in vanilla SNNs, which significantly reduces the information loss caused by the spike quantization (see Table~\ref{Table:QE_Value}). It is evident from a comparison of Figure~\ref{fig:MPD_analysis}\textbf{a} and \textbf{b} that each channel's MPD grows thinner (the variance becomes smaller). In this work, the reduced variance implies that the edge information in the spike feature is clearer from the perspective of feature visualization. By contrast, from the perspective of QE, it implies that the information loss becomes less.


\section{Conclusions}
In this work, three key questions are exploited to analyze the redundancy of SNNs, which are usually ignored in other prior works. To answer these questions, we present a new perspective on the relationship between the spatio-temporal dynamics and the spike firing. These findings inspired us to develop a simple yet efficient advanced spatial attention module for SNNs, which harnesses the inherent redundancy in SNNs by optimizing the membrane potential distribution. Experimental results and analysis show that the proposed method can greatly reduce the spike firing and further improve performance. \rone{The new insight onto SNN redundancy not only reveals the unique advantages of spike-based neuromorphic computing in terms of bio-plausibility, but may also bring some interesting enlightenment to the following-up research on efficient SNNs.}

\section*{Acknowledgement}
This work was partially supported by National Science Foundation for Distinguished Young Scholars (62325603), and National Natural Science Foundation of China (62236009, U22A20103), and Beijing Natural Science Foundation for Distinguished Young Scholars (JQ21015).

{\small
\bibliographystyle{ieee_fullname}
\bibliography{egbib}
}

\end{document}